\documentclass[sigconf]{acmart}

\AtBeginDocument{%
  \providecommand\BibTeX{{%
    \normalfont B\kern-0.5em{\scshape i\kern-0.25em b}\kern-0.8em\TeX}}}

\usepackage{xspace}
\usepackage{xcolor}
\usepackage{enumitem}
\usepackage{bm}
\usepackage{graphicx}
\usepackage{subcaption}
\usepackage{amsmath}
\usepackage{url}
\usepackage{array,multirow}
\usepackage[utf8]{inputenc}
\usepackage[english]{babel}

\usepackage{mathtools}
\DeclarePairedDelimiter{\ceil}{\lceil}{\rceil}

\usepackage{amsthm}
\theoremstyle{plain}

\theoremstyle{remark}

\usepackage{adjustbox}
\usepackage[ruled,vlined]{algorithm2e} 

\newcommand{\modelname}{\texttt{SimpDOM}\xspace}
\newcommand{\modelfullname}{\underline{Simp}lified \underline{DOM} Trees for Attribute Extraction\xspace}

\DeclareMathOperator*{\argmax}{arg\,max}

\usepackage{array,booktabs,ragged2e}
\newcolumntype{R}[1]{>{\RaggedLeft\arraybackslash}p{#1}}

\setcopyright{iw3c2w3}

\begin{document}



\title{Simplified DOM Trees for Transferable Attribute Extraction from the Web}





\author{Yichao Zhou}
\authornote{The work was done while BYL was a research intern at Google AI.}
\affiliation{%
  \institution{University of California, Los Angeles}
  \city{Los Angeles}
  \state{California}
  \country{USA}
}
\email{yz@cs.ucla.edu}

\author{Ying Sheng}
\affiliation{%
  \institution{Google}
  \city{Mountain View}
  \state{California} 
  \country{USA}
}
\email{yingsheng@google.com}

\author{Nguyen Vo}
\affiliation{%
  \institution{Google}
  \city{Mountain View}
  \state{California} 
  \country{USA}
}
\email{nguyenvo@google.com}

\author{Nick Edmonds}
\affiliation{%
  \institution{Google}
  \city{Mountain View}
  \state{California} 
  \country{USA}
}
\email{nge@google.com}

\author{Sandeep Tata}
\affiliation{%
  \institution{Google}
  \city{Mountain View}
  \state{California} 
  \country{USA}
}
\email{tata@google.com}

\begin{abstract}

There has been a steady need to precisely extract structured knowledge from the web (i.e. HTML documents). Given a web page, extracting a structured object along with various attributes of interest (e.g. price, publisher, author, and genre for a book) can facilitate a variety of downstream applications such as large-scale knowledge base construction, e-commerce product search, and personalized recommendation. 
Considering each web page is rendered from an HTML DOM tree, existing approaches formulate the problem as a DOM tree node tagging task. However, they either rely on computationally expensive visual feature engineering or are incapable of modeling the relationship among the tree nodes. 
In this paper, we propose a novel transferable method, \modelfullname (\modelname), to tackle the problem by efficiently retrieving useful context for each node by leveraging the tree structure. We study two challenging experimental settings: (i) intra-vertical few-shot extraction, and (ii) cross-vertical few-shot extraction with out-of-domain knowledge, to evaluate our approach.
Extensive experiments on the SWDE public dataset show that \modelname outperforms the state-of-the-art (SOTA) method by 1.44\% on the F1 score. We also find that utilizing knowledge from a different vertical (cross-vertical extraction) is surprisingly useful and helps beat the SOTA by a further 1.37\%.

\end{abstract}


\copyrightyear{2021}
\acmYear{2021}
\acmConference[WWW '21]{Proceedings of The Web Conference 2021}{April 19-23, 2021 }{Ljubljana, Slovenia}
\acmBooktitle{Proceedings of The Web Conference 2021 (WWW '21), April 19-23, 2021, Ljubljana, Slovenia}
\acmPrice{}



\maketitle

\section{Introduction} \label{sec:introduction}

\begin{figure}[t]
\centering
  \includegraphics[width=\columnwidth]{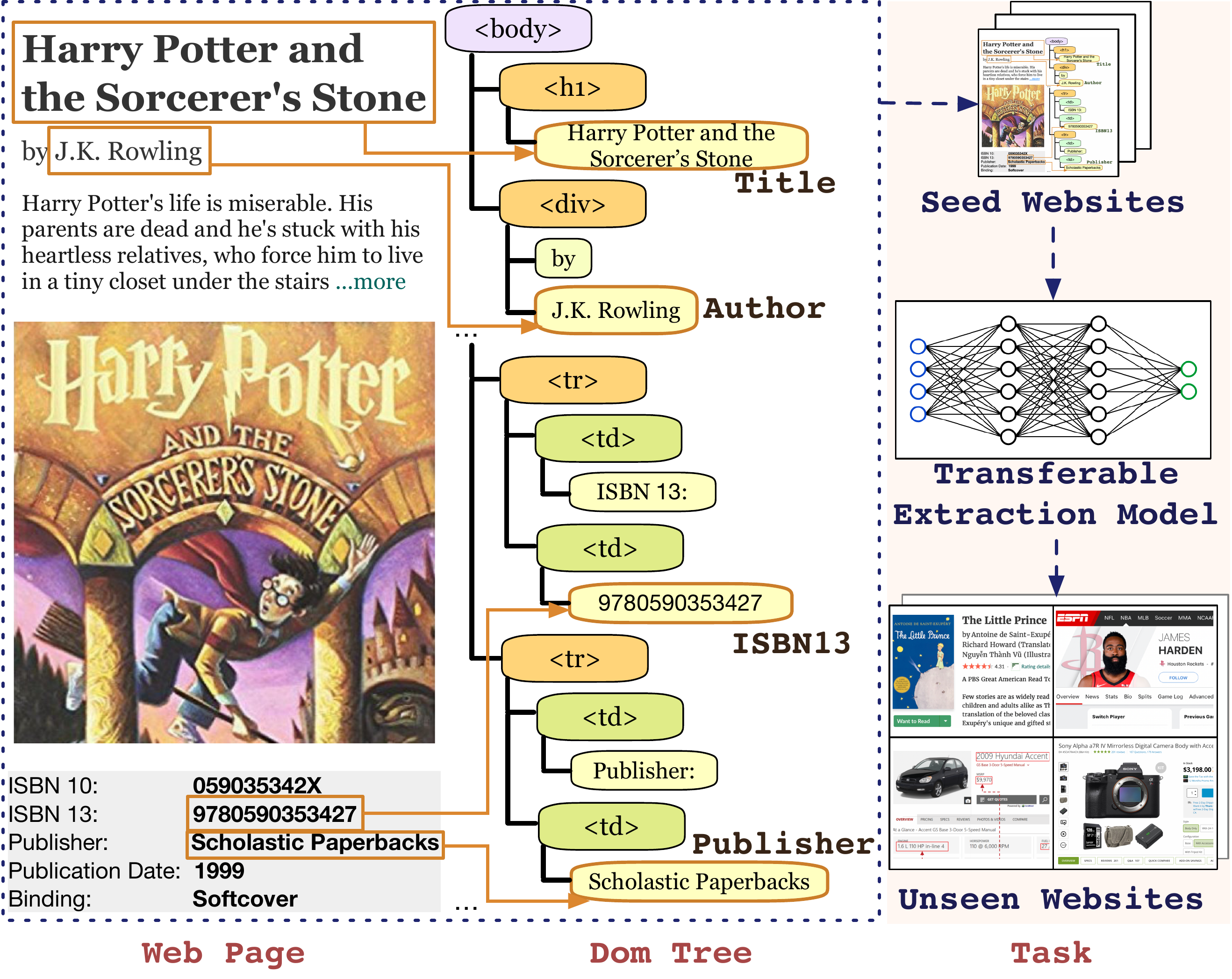}
  \caption{Learning a transferable model based on HTML DOM trees to extract attributes from unseen websites of various verticals. Note that every web page is rendered from a Document Object Model (DOM) Tree. All the attributes are located in the leaf nodes of the trees.}
  \label{fig:pull_figure}
\end{figure}

As the world wide web explosively grows nowadays, there has been a perennial need to automate the translation of web pages into structured knowledge~\cite{chang2006survey,hao2011one}. 
Attribute extraction systems recognize attributes of interest from web pages. For example, collecting book authors can facilitate the user's faceted search by allowing users to narrow down the search results with a filter on the book's attribute. Attribute extraction as well enables various downstream applications including large-scale knowledge base/graph construction~\cite{dong2014knowledge,wu2018fonduer}, e-commerce product search~\cite{bing2016unsupervised,hao2011one}, and personalized recommendation~\cite{wang2019multi}. 
However, the semi-structured data format, noisy page contents, and multifarious page layouts all make it a non-trivial task to extract attributes, compared to the unstructured texts, which can be easily modeled as a sequence~\cite{lockard2018ceres}. 

Take Figure~\ref{fig:pull_figure} as an example. The web page on the left-hand-side is a partial screenshot from a bookstore website. The web page is rendered to display in a browser based on the source data, a Document Object Model (DOM) tree~\cite{gupta2003dom} (a corresponding subtree is shown in the middle of Figure~\ref{fig:pull_figure}). In this paper, our goal is to extract attributes of interest such as \{\textit{title, author, isbn13, publisher}\} from the detail pages of various websites. A detail page denotes a page that corresponds to a single data record~\cite{carlson2008bootstrapping}, like a book in a bookstore or an NBA player on a sports website. 

Traditional solutions rely on the fact that many websites are created by templates such as Wrapper Induction~\cite{kushmerick1997wrapper,muslea1999hierarchical,azir2017wrapper}. Some unsupervised methods~\cite{chang2001iepad,zhai2005web} avoid the use of templates and can automatically extract attributes, but neglect the semantics of attribute values.
Thus considerable human efforts are required for either periodically updating templates or annotating unseen websites. In this work, we aim to build a novel transferable model to reduce expensive human efforts and to extract attributes from unseen websites of various verticals. 

Some recent work~\cite{hao2011one,lockard2020zeroshotceres} explores visual patterns of each node such as its bounding box coordinates and surrounding nodes on the web page. However, achieving these features requires a computationally expensive rendering process and extra memory space to save the necessary images, CSS, and JavaScript files that could easily be out-of-date. \texttt{FreeDOM}~\cite{lin2020freedom} avoids rendering-based features and models the pairwise node relationship with node-level feature representations that are learned separately. Nevertheless, it is inconvenient to deploy such a two-stage model in practice. The rich DOM tree-level contexts are neglected by this method as well. In this paper, we propose a novel single-stage approach, \modelfullname (\modelname), that does not require visual features, instead relying on a careful construction of the context of a node in the DOM tree that generalizes well to unseen websites in the domain\footnote{We use domain and vertical interchangeable in this paper.} as well as to websites in other domains.

\begin{figure}[t]
\centering
  \includegraphics[width=\columnwidth]{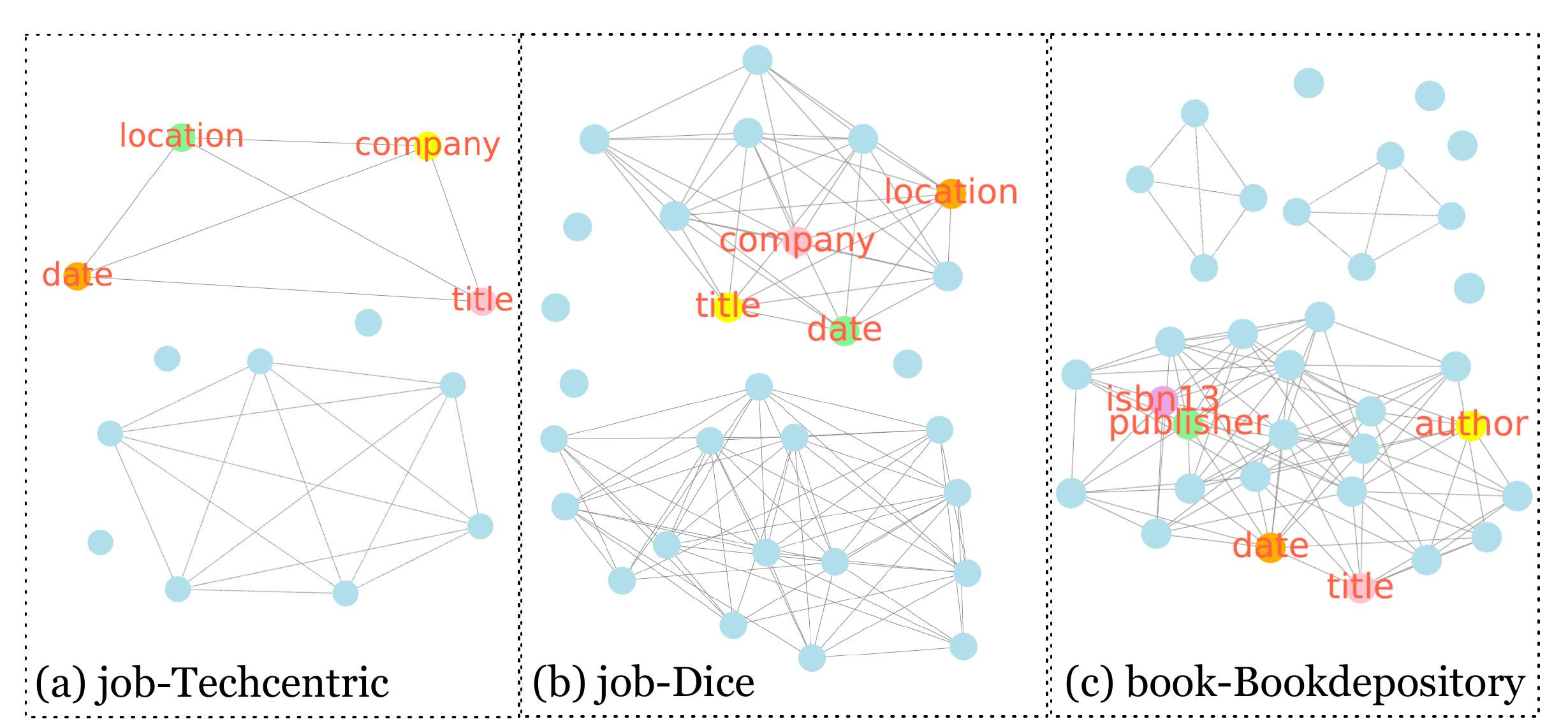}
  \caption{Graph visualization of the DOM node neighborhood. Each node is linked to its close neighbors depending on the DOM tree structures. 
  A site-invariant feature can be concluded that nodes containing attribute values are usually close to each other in the DOM trees. }
  \label{fig:graph}
\end{figure}

Specifically, \modelname builds a rich representation for each node by focusing on its contextual features. Then a node classification is conducted to decide which attribute type it belongs to.
For instance in Figure~\ref{fig:pull_figure}, we notice that the closest text node to ``J. K. Rowling'' contains information ``by'' which means ``J. K. Rowling'' is likely to be the \textit{author} of this book. 
We also notice important attribute values are usually clustered together to draw readers' attention, like \textit{isbn13}, \textit{publisher}, and \textit{publication date} appear right in the same table.
In short, the contexts in a simplified neighborhood can provide website-invariant features such as some semantically informative expressions and vertical-invariant clues such as the co-occurrence of multiple attribute values. 
We visualize the neighboring relationship of DOM nodes for three websites from two verticals in figure~\ref{fig:graph}. Obviously, the nodes that contain attribute values are always close to each other in the DOM trees. 

We consider two challenging experimental scenarios in this paper, (i) intra-vertical few-shot extraction, where we learn a model with a few labeled seed websites and predict on other unseen websites from the same vertical; (ii) cross-vertical few-shot extraction with out-of-domain knowledge, where we train the model with all the websites from an out-of-domain vertical \textit{A}, then finetune this model with a few seed sites from vertical \textit{B}, and finally test it on other unseen websites from vertical \textit{B}. The first scenario tests the transferability among websites from the same vertical while the second assesses the effectiveness of cross-domain knowledge.

Overall, our paper describes the following contributions:
\begin{itemize}
    \item To the best of our knowledge, this is the first work that efficiently extracts each node's informative contexts from the DOM trees to tackle the attribute extraction task. 
    \item We are also the first to study the transferable representations for the cross-vertical few-shot attribute extraction scenario. 
    \item Extensive experiments on the public dataset, SWDE~\cite{hao2011one}, show that \modelname significantly outperforms the SOTA method by 1.44\% on the F1 score, and the out-of-domain knowledge helps beat the SOTA by a further 1.37\%. 
    \item We will open-source our implementations to provide a testbed and facilitate future research in this direction.
\end{itemize}
\section{Problem Formulation and Approach} \label{sec:method}
In this section, we formally define the problem and introduce the outline of our proposed method, \modelname.
\begin{figure*}[t]
\centering
\begin{minipage}[t]{0.63\textwidth}
\centering
  \includegraphics[width=\linewidth]{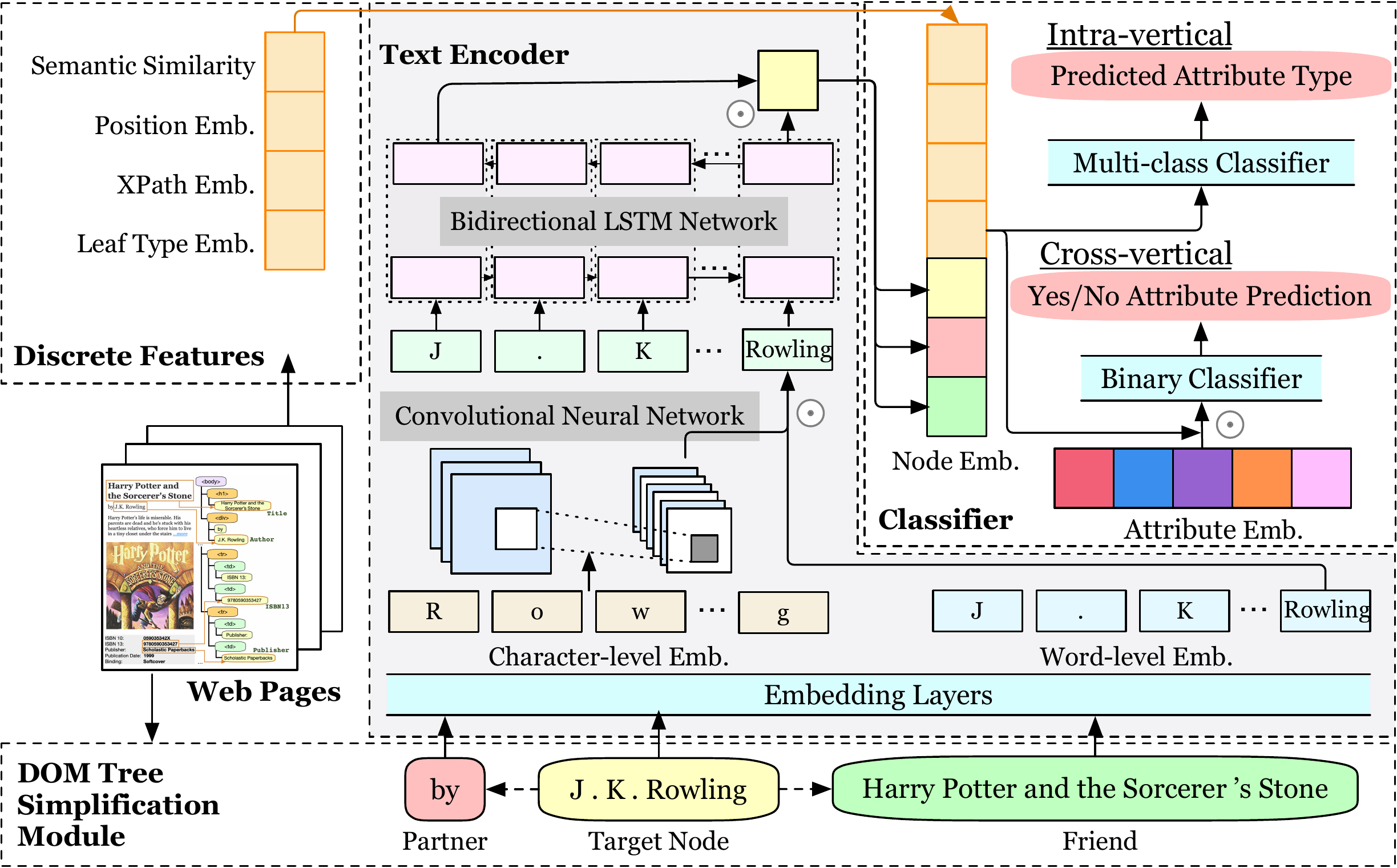}
  \caption{The overall architecture of \modelname. Nodes' textual features are encoded by LSTM and CNN at the word-level and character-level respectively. A set of discrete features are built from the DOM trees including leaf type, XPath, and the relative position of each node. [$\cdot \odot \cdot$] denotes vector concatenation.}
  \label{fig:model}
\end{minipage}\hfill
\begin{minipage}[t]{0.35\linewidth}
\centering
  \includegraphics[width=\linewidth]{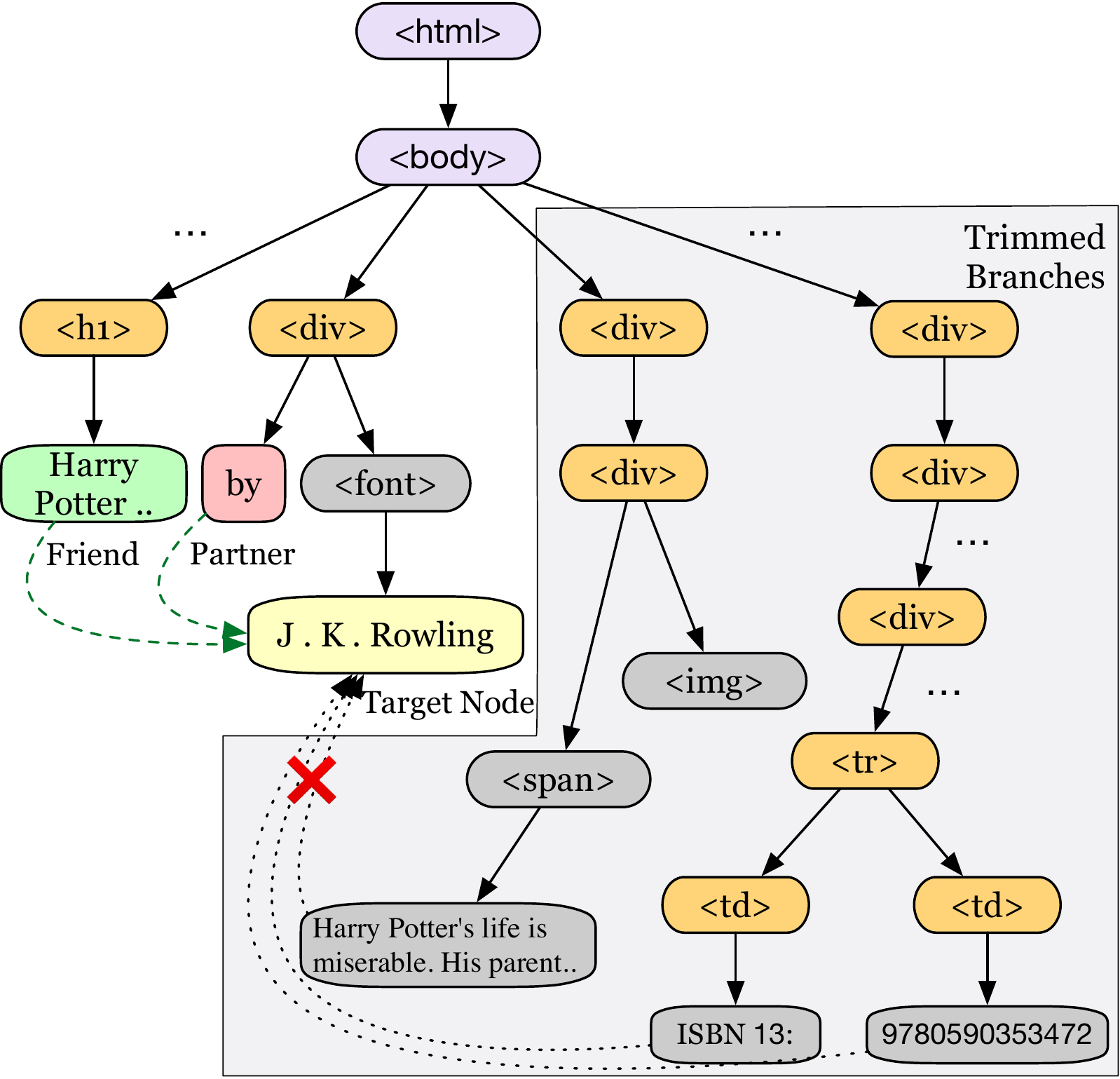}
  \caption{The DOM Tree Simplification Module extracts the partner (by) and friends (Harry Potter and the Sorcerer's Stone) for each node (J. K. Rowling) by trimming unrelated branches. }
  \label{fig:simpdom}
\end{minipage}
\end{figure*}

\subsection{Few-shot attribute extraction from semi-structured websites}
We tackle the problem of extracting attributes from unseen semi-structured websites.
Each \textit{vertical} $V$ has a set of websites. Each \textit{website} $W$ is composed of a collection of \textit{detailed pages} which share a similar template.
Each page has a DOM tree $T$ which contains a variable node set $X$ and a fixed node set $Y$ where the text contents are stored, and also a set of non-text nodes $Z$. Fixed nodes remain the same across different detailed pages on the same website while variable nodes may contain different contents. 

\noindent \textbf{Attribute Extraction.}
The goal of attribute extraction is to extract a possible value for each attribute type from the DOM tree nodes. We narrow down the search range to variable nodes because the attribute values should vary in different detailed pages. We formulate the attribute extraction as a node tagging task. Given a detailed page $p$ with a set of variable nodes $X$, we aim to learn a model to classify each node $x \in X$ into one of the pre-defined vertical-specific types (e.g. \textit{title, author, isbn13, publisher}) or \textit{none} representing that this node does not contain any attribute values. We assume that one node can correspond to at most one pre-defined attribute type~\cite{hao2011one}.

\noindent \textbf{Few-shot Intra-vertical Extraction.}
Given a set of annotated seed websites $\{W_1^a, W_2^a, ..., W_i^a\}$ from vertical $V$, we aim to learn a transferable model $\mathcal{M}$ to extract attributes from a larger set of unseen websites $\{W_1^u, W_2^u, ..., W_j^u\}$ from the same vertical. 

\noindent \textbf{Few-shot Cross-vertical Extraction.}
In this scenario, we leverage a set of annotated out-of-domain websites from vertical $V_1$ to learn a transferable extraction model $\mathcal{M}_o$ and fine-tune the model with seed websites $\{W_1^a, W_2^a, ..., W_i^a\}$ from vertical $V_2$. Finally, we extract attributes from unseen websites $\{W_1^u, W_2^u, ..., W_j^u\}$ of $V_2$.

\subsection{Approach Overview}

Figure~\ref{fig:model} shows the overall framework of the proposed \modelname model for the few-shot attribute extraction task.
We firstly simplify the DOM trees to extract context features for each node. All the textual features are then fed into a text encoder to generate a dense semantic embedding. We find that extra discrete features built from markup information such as XPath and leaf node type can result in a better node representation. We also add the relative position of each node as a global feature for the extraction task. The combined node embedding is used for predicting the type of node. In the intra-vertical scenario, we directly apply a multi-class classifier to the node embedding and output the attribute type probability distribution. 
In the cross-vertical scenario, the attribute sets differ from vertical to vertical. Therefore, we have to alter the inference strategy to binary classification to achieve a matching probability for each attribute type. Then, we select the attribute with the highest probability as the prediction.

\section{Node Encoder and Classifier}
The node encoder consists of three components: DOM tree simplification module, text encoder, and discrete feature module. 
\subsection{DOM Tree Simplification Module}

\begin{algorithm}[t]
\SetAlgoLined
\textbf{Input:} DOM tree $T$\, variable node set $X$\, constant $K$;
\textbf{Output:} Dictionaries $D_p$ and $D_f$ where each key is $x \in X$ and the values are its corresponding partner and friend set, respectively\;

    Initialize $D_d, D_p, D_f$ as three empty dictionaries\; 
    \For{each variable node $x \in X$}{
    Get the node's XPath $P_{x}$ from $T$\;
    Generate the node's ancestor set $ANC_{x}$ from $P_{x}$ and mark the ordered closest $K$ ancestors as $ANC_{x}^{K}$\;
        \For{each $anc$ in $ANC_x^K$}{
            Add $x$ to $D_d[anc]$\;
        }
    }
    \For{each variable node $x \in X$}{
        \For{each $anc \in ANC_x^K$}{
            $DESC \leftarrow D_d[anc] \backslash \{x\}$\;
            \uIf{there exists only one node $x'$ in $DESC$ and both $D_p[x], D_f[x]$ are empty}
            {Add $x'$ to $D_p[x]$\;}
            $D_f[x] \leftarrow D_f[x] \cup DESC$\;
        }
    }
\caption{Function $\mathcal{F}$ for DOM Tree Simplification and Friend Circle Extraction.}
\label{algo:simpdom}
\end{algorithm}

In this module, we simplify the DOM tree to extract the contexts for each variable node $x$, namely its friend circle features, which are composed of the node's \textit{partner} and \textit{friends}. 
The whole DOM tree is a collection of nodes that originate from a unique starting node called the \textit{root}. The set of nodes $A$ on the path from \textit{root} to node $x$ (not including $x$) are ancestors of node $x$. 
The \textit{friends} of $x$ denotes a set of text nodes $X^F \subseteq X \cup Y \backslash \{x\}$. For each $x^f \in X^F$, the distances from both $x^f$ and $x$ to their lowest common ancestor $a \in A$ should be no more than constant $N$. We compute the distance by counting the number of edges on the path.
The \textit{partner} $x^p$ of $x$ is a special \textit{friend} node for which $x$ and $x^p$ are the only two text nodes in the tree that originates from their lowest common ancestor.
Note that each node has at most one \textit{partner} in the DOM tree while it could have zero or multiple \textit{friends}. Usually, \textit{partner} $x^p$ is the closest \textit{friend} to $x$ in the DOM tree.

As function $\mathcal{F}$ described in Algorithm~\ref{algo:simpdom}, for each variable node $x \in X$, we decode its XPath information to record the $K$ closest ancestors of $x$. For instance, if the XPath of $x$ is ``/body/tr/td/'', we consider both ``/body/tr/'' and ``/body/'' as the ancestor of $x$. 
Reversely, we can easily obtain all the descendants of each ancestor node to composite the candidate set for retrieving the partner and friends.
By limiting the size of $K$, we can narrow down the search area in the tree such that the noisy textual features from distant branches can be efficiently trimmed, as shown in Figure~\ref{fig:simpdom}.

\begin{figure}[h]
\centering
  \includegraphics[width=.75\linewidth]{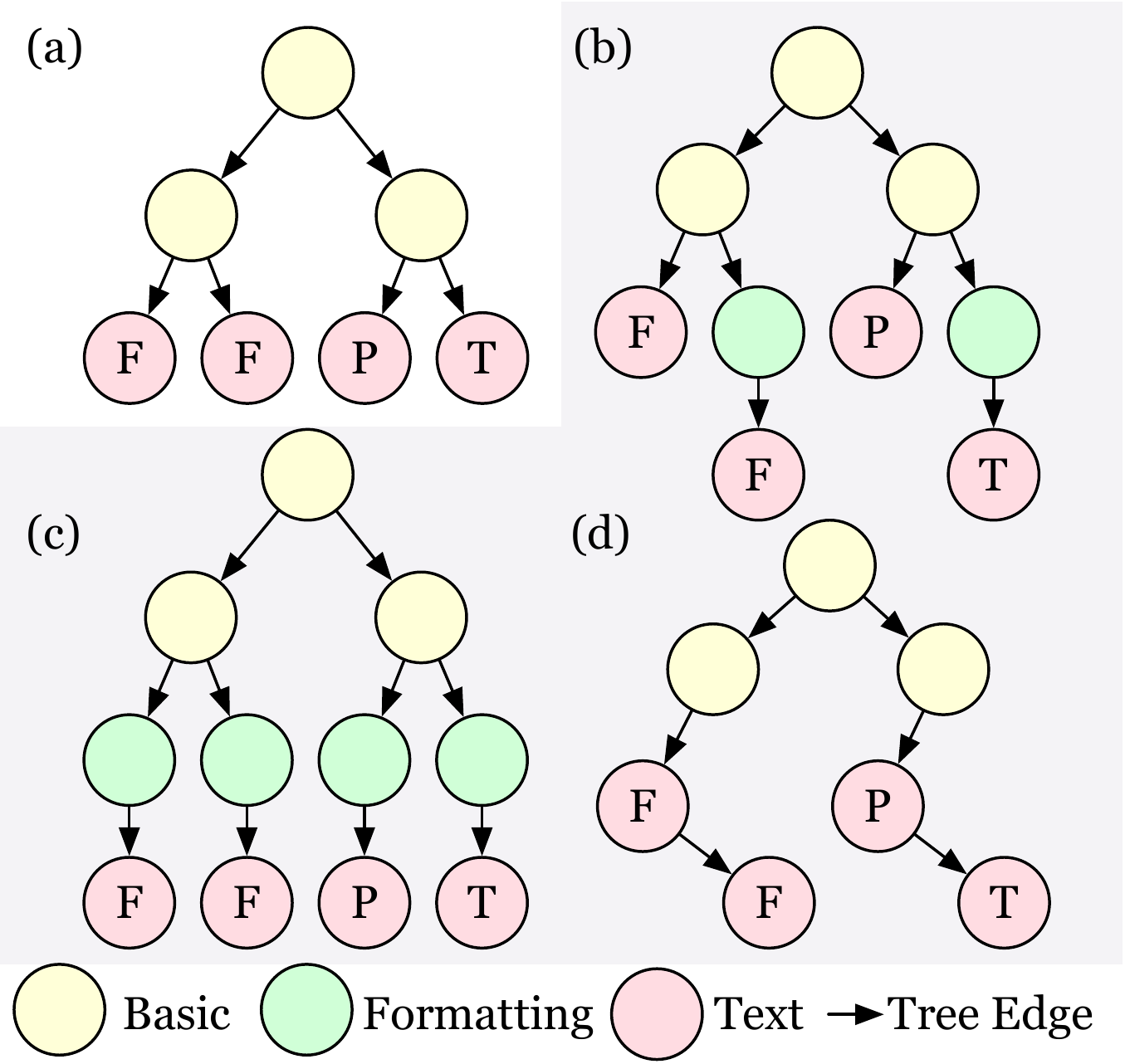}
  \caption{Subtree skeletons of web page DOMs including a common structure (a) and its three possible variants (b), (c) and (d). ``Basic'' denotes a set of basic HTML element tags, while ``Formatting'' represents some formatting and style tags such as <strong> and <font>. The ``Text'' node has text information. We aim to simplify all possible variants to (a), in order to efficiently extract the partner (marked as P) and friends (F) for each target node (T).}
  \label{fig:trees}
\end{figure}

In the extraction process, we keep all the basic HTML element tags like <tr> and <td> while remove the formatting and style tags such as <strong> and <font>\footnote{We refer to the HTML tag categories in \url{https://www.w3schools.com/TAGS/ref_byfunc.asp}.}. In Figure~\ref{fig:trees}, we plot a common sub tree structure (a) and its three possible variants (b,c,d). With Algorithm~\ref{algo:simpdom}, we can simplify and normalize the three variants to (a) in order to extract the friend circle features.

With partner and friends extracted from the DOM tree for each node $x$, we feed the three sets of textual features separately into the text encoder as described in section~\ref{sec:encoder} to generate three representations $e_x, e_p,$ and $e_f$ which are all $d_w$-dimensional vectors.
We derive the joint semantic embedding $e_s$ by simply concatenating the three representations as follows:
$$e_s = \left[e_x; e_p; e_f\right].$$
Note that the joint embedding is a $3d_w$-dimensional vector.

\subsection{Text Encoder}\label{sec:encoder}
Node $x$ contains a sequence of text $S_1 = [w_1, w_2, ..., w_{L1}]$, where $w_i\in \mathcal{W}$ and $L1$ denotes the word sequence length. We can easily split each word into a sequence of characters $S_2 = [c_1, c_2, ..., c_{L2}]$, where $c_i\in \mathcal{C}$ and $L2$ is the character sequence length. $\mathcal{W}$ and $\mathcal{C}$ are vocabularies of words and characters. 
We employ a hierarchical LSTM-CNN text encoder to encode the character-level and word-level features. 

We notice that the attribute values usually contain useful morphological patterns in the character-level semantics~\cite{lin2020freedom}. For example, ($aa$'$bb$ ft) and ($aa$-$bb$ ft) are two common patterns of \textit{height} attribute in the \textit{nbaplayer} vertical. Their character-level representation can be very essential. Therefore, we leverage a Convolutional Neural Network to encode the character-level embeddings (dimension $d_c$) of each word $w$, resulting in $h^c_w$. We simply concatenate $h^c_w$ with its word-level representation $g_w$ retrieved from external pretrained word embeddings: $h_w = \left[g_w; h^c_w\right]$.  

The LSTM~\cite{hochreiter1997long} has been widely used as the unit of Recurrent Neural Network for learning the latent representation of sequence data~\cite{liu2016recurrent}. 
Therefore, we feed the latent word representations $\left[h_{w_1}, h_{w_2}, ..., h_{w_{L1}}\right]$ into a bi-directional LSTM network, resulting in $e_x = \left[h_w^{forward}; h_w^{backward}\right]$.

Similarly, we can achieve the semantic representations for the node's partner and friends, $e_p$ and $e_f$.

\subsection{Discrete Feature Module}
\noindent\textbf{Xpath embeddings.}
Markup features such as XPath can be very useful for node tagging. An XPath of a DOM node ``/html/body/tr/td/'' can be seen as a sequence of HTML tags $[$<html>, <body>, <tr>, <td>$]$. 
We learn a separate bi-directional LSTM to get the dense representation $e_{xpath}$ of dimension $d_{xpath}$ for each XPath sequence such that it can make use of all the meaningful tags in the sequence. 

\noindent\textbf{Leaf node type embeddings.}
The tag type of the DOM leaf node such as ``<h1>'' can also be meaningful. ``<h1>'' means the node is likely to be the title of the page, highly correlating with the \textit{name} of a \textit{nbaplayer} or the \textit{title} of a \textit{book}. We collect the vocabulary set of the HTML tags and randomly initialize an embedding $e_{leaf}$ of dimension $d_{leaf}$ for each of them.

\noindent\textbf{Position embeddings.}
We also leverage the relative position of each node $x$ as a discrete feature. This global information can benefit the task. For example in the \textit{auto} vertical, the \textit{model} usually lies on the top of the page. We apply depth-first-search to traverse the tree and get the occurrence position $pos_x$ of each node. Then we compute its relative position via $\ceil{\frac{pos_x}{\max_x\{pos_x\}}}$. Similarly, a random embedding $e_{pos}$ of dimension $d_{pos}$ is initialized for each position.

\noindent\textbf{Semantic similarity.}
We notice the for each node $x$ the text in the partner node $x^p$ can help determine $x$'s attribute type and modeling the semantic relation between the text in $x^p$ and the attribute types allows us to best leverage this data. Specifically, we compute the \textit{cosine similarity}\footnote{We compute the scores via \textit{cosine\_similarity} $(e_p, e_{a_i})= \frac{e_p\cdot e_{a_i}}{|e_p||e_{a_i}|}$.} between the partner embedding $e_p$ and each attribute embedding $e_{a_i}$ to model their semantic relations, which results in a semantic similarity vector $e_{cos}$ of dimension $M$, where $M$ denotes the number of pre-defined attribute types. 

Upon achieving these discrete features, we concatenate them into a vector $e_d=\left[ e_{xpath}; e_{leaf}; e_{pos}; e_{cos}\right]$ of dimension $d_{xpath} $+$ d_{leaf} $+ $ d_{pos} $+$ M$.

\subsection{Inference and Optimization}
Under the intra-vertical scenario, the node embedding is connected to a multi-layer perceptron (MLP) for multi-class classification, as illustrated below:
$$e_n = \left[e_s; e_d\right]$$
$$\text{\textbf{h}} = \text{MLP}(e_n), \text{\textbf{\textbf{h}}}\in \mathbb{R}^{M+1}.$$
where $M+1$ denotes the number of pre-defined attribute types plus a \textit{none} type.

Under the cross-vertical scenario, we notice each vertical has a different attribute set. The MLP layer for multi-class classification can no longer be reused for different verticals which have different sizes of attribute sets. Therefore, we alter the inference strategy to binary classification. We individually concatenate the node embedding $e_{n}$ to each attribute embedding $e_{a_i}$ of dimension $d_a$ which is randomly initialized. We then connect it to a separate MLP and compute a score \textbf{h}$_i$ for each attribute type:
$$e_{b_i} = \left[e_n; e_{a_i}\right], 1 \leq i \leq M+1$$
$$\text{\textbf{h}}_i = \text{MLP}(e_{b_i}), \text{\textbf{h}}_i\in \mathbb{R}$$

Under both scenarios, we lastly apply the \textit{softmax} function to normalize \textbf{h} and select the largest as the prediction $\hat{\text{\textbf{y}}}$:
$$\text{\textbf{p}}_i =  \frac{\text{e}^{\text{\textbf{h}}_i}}{\sum^{M+1}_{j=1} \text{e}^{\text{\textbf{h}}_j}}; \hat{\text{\textbf{y}}} = \argmax_i \text{\textbf{p}}_i.$$

The loss function optimizes the cross-entropy between the true labels $\text{\textbf{y}}$ and the normalized probabilistic scores $\text{\textbf{p}}$.
$$\text{loss} = -\sum_{n=1}^{|X|} \sum_{m=1}^{M+1} \text{\textbf{y}}_{m,n} \log \text{\textbf{p}}_{m,n} $$

\section{Experiments}\label{sec:results}
In this section, we firstly introduce the dataset and evaluation metrics. We also explain the implementation details to guarantee the reproducibility of our method. Then, a collection of baseline models are introduced to compare with our model under the intra-vertical few-shot extraction scenario. We also conduct a series of ablation studies to answer the following questions: (i) \textit{What are the contributions from each set of features?} (ii) \textit{Will sequence modeling work well on DOM tree nodes?} (iii) \textit{What are the performances of using different word embedding strategies?} Lastly, we evaluate the effectiveness of the out-of-domain knowledge under the cross-vertical few-shot extraction scenario.

\subsection{Dataset}
We rely on a public data set, SWDE~\cite{hao2011one} that consists of more than 124,000 web pages from 80 websites of 8 verticals to train and evaluate the proposed model. Detailed statistics are shown in Table~\ref{tab:dataset}. Each vertical consists of 10 websites and contains 3 to 5 attributes of interest. We notice \textit{book} and \textit{job} have the most variable nodes on average which is roughly three times the nodes in vertical \textit{auto} and \textit{university}.

\begin{table}[h]
    \centering
    \resizebox{\linewidth}{!}{
    \begin{tabular}{|c|c|c|c|c|c|}
    \hline
        Vertical & \#Sites & \#Pages & \#Var. Nodes & Attributes \\
        \hline\hline
        auto & 10 & 17,923 & 130.1 & model, price, engine, fuel \\
        book & 10 & 20,000 & 476.8 & title, author, isbn13, pub, date \\
        camera & 10 & 5,258 & 351.8 & model, price, manufacturer \\
        job & 10 & 20,000 & 374.7 & title, company, location, date \\
        movie & 10 & 20,000 & 284.6 & title, director, genre, mpaa \\
        nbaplayer & 10 & 4,405 & 321.5 & name, team, height, weight \\
        restaurant & 10 & 20,000 & 267.4 & name, address, phone, cuisine \\
        university & 10 & 16,705 & 186.2 & name, phone, website, type \\
    \hline
    \end{tabular}
    }
    \vspace{10pt}
    \caption{SDWE Dataset Statistics}
    \vspace{-15pt}
    \label{tab:dataset}
\end{table}

In the intra-vertical few-shot experiments, we follow the settings in FreeDOM~\cite{lin2020freedom} to randomly select $k$ seed websites as the training data and use the remaining $10-k$ websites as the test set. Note that in this few-shot extraction task, none of the pages in the $10-k$ websites have been visited in the training phase. This setting is abstracted from the real application scenario where only a small set of labeled data is provided for specific websites and we aim to infer the attributes on a much larger unseen website set.

In the cross-vertical few-shot experiments, we leverage one vertical as the out-of-domain knowledge to train a model. Then we conduct the same intra-vertical extraction experiments by loading the checkpoints from the pretrained model for parameter initialization. We create this experimental setting to enable a broader knowledge transfer across various verticals, which can tackle the scenario where the domain of the existing annotation is inconsistent with the unseen websites. 

\subsection{Evaluation Metrics}
\begin{figure*}[h]
\centering
  \includegraphics[width=\linewidth]{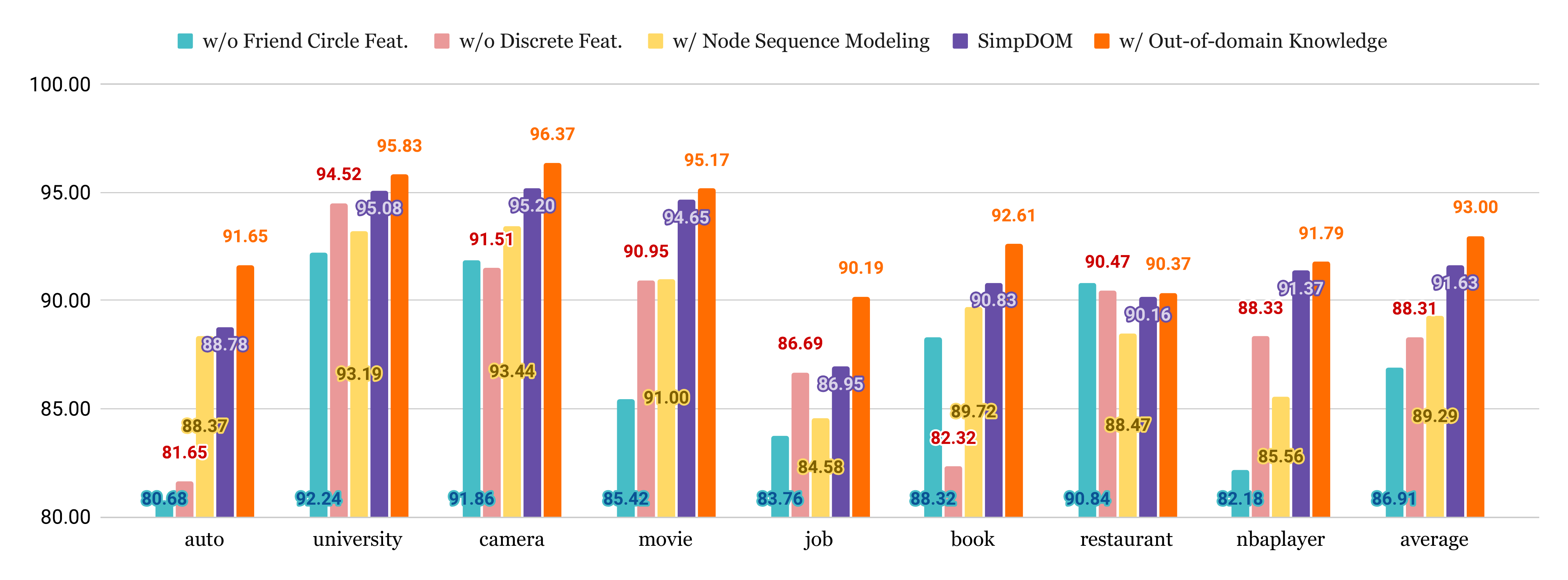}
  \caption{Ablation study results that demonstrate the contribution from different features and modules. We conclude that both friend circle features and discrete features improve the extraction performance while adding a sequence modeling module harms the performance dramatically. With the out-of-domain knowledge from a second vertical, the model can do better for each of them. We set $k=3$ here. Similar results can be achieved with other $k$'s. }
  \label{fig:ablation}
\end{figure*}

\begin{figure*}[h]
\centering
  \includegraphics[width=.85\linewidth]{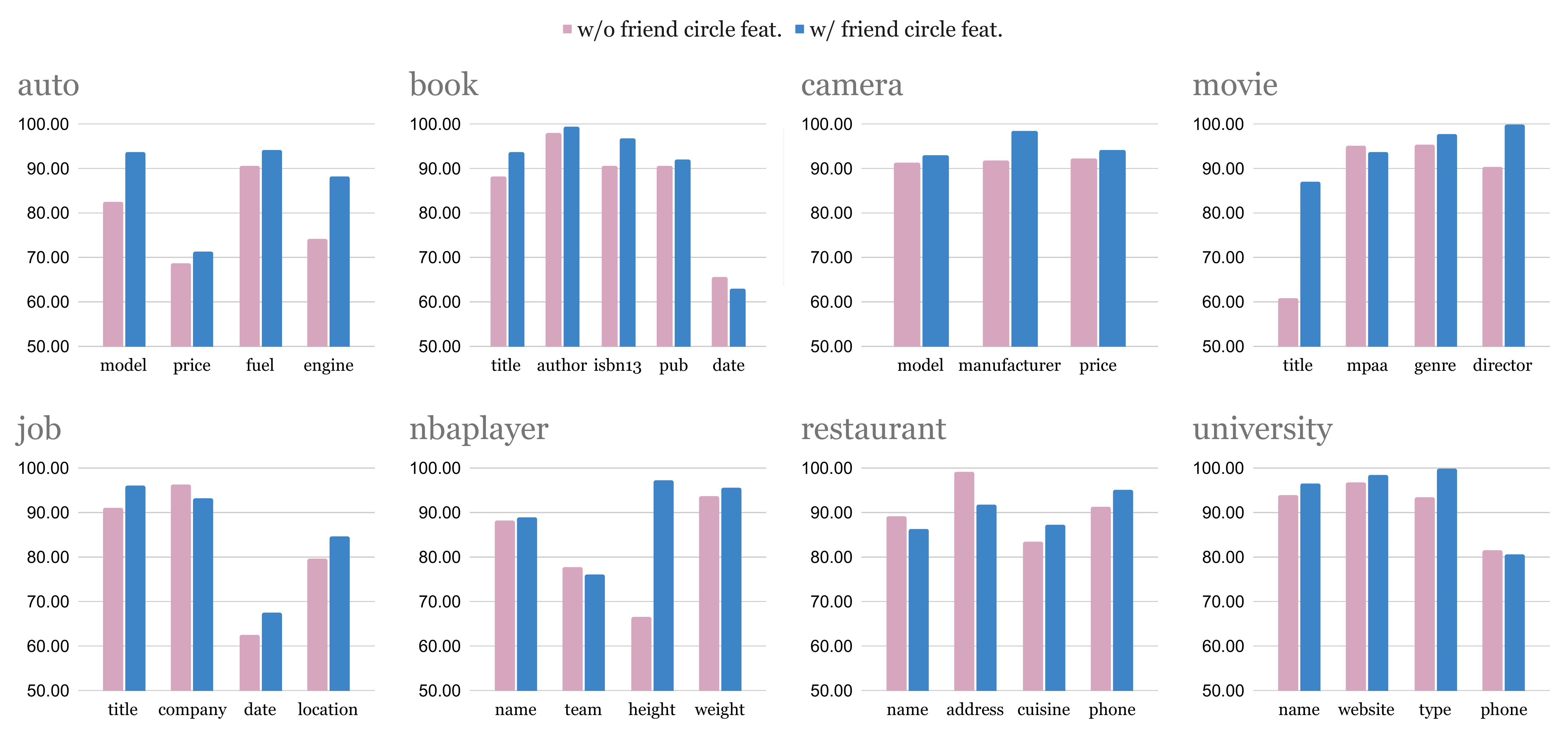}
  \caption{Per-attribute F1 performance comparisons between \modelname w/ and w/o friend circle features. We set $k=3$ here. Attributes like \textit{height} in \textit{nbaplayer} and \textit{title} in \textit{movie} get the largest performance lifts.}
  \label{fig:attributes}
\end{figure*}

We evaluate the extraction performance by page-level F1 scores, following the evaluation metrics from SWDE and FreeDOM~\cite{lin2020freedom,hao2011one}. Page-level F1 score is the harmonic mean of extraction precision and recall in each page. Specifically, we evaluate the predicted attribute values with the true values for each detailed page.
We compute an average F1 score over all the verticals (Table~\ref{tab:main}) to compare with the baselines.
We also compute the average F1 score for each vertical (Figure~\ref{fig:ablation}) and each attribute (Figure~\ref{fig:attributes}) for detailed analysis. 

\subsection{Implementation details}
For data pre-processing, we use open-source LXML library\footnote{\url{https://lxml.de/}} to
process each page for obtaining the DOM tree structures. 
Then, we follow the simple heuristic used in \cite{lin2020freedom} to filter nodes whose values are constant in all pages of a website, thus most of the noisy page-invariant textual nodes such as the footer and navigation contents are removed and the experiments are significantly accelerated in terms of the training speed. We use GloVe pretrained representations~\cite{pennington2014glove} to initialize our word embeddings. Other representations such as character embeddings and attribute embeddings are all randomly initialized. We also cut off every node's text when it has more than $15$ words. We set both maximum edge number $N$ and maximum ancestor number $K$ as 5 for extracting friend circle features and only keep the closest $10$ friends for each DOM tree node by comparing their relative positions on the web page.

We conduct a grid search for all the hyper-parameters. We use $100$ for both word embedding size $d_{w}$ and character embedding size $d_{c}$. We select 
$d_{path}$, $d_{leaf}$, $d_{pos}$ as $30, 30, 20$, respectively. For the CNN network, we use $50$ filters and $3$ as kernel size. For the LSTM network, we set the hidden layer size as $100$. The model is implemented in Tensorflow. We train the model with epoch number $15$ and a batch size $32$. We apply a dropout mechanism following the MLP layer to avoid over-fitting issues. The dropout rate is $0.3$. We use Adam as the optimizer where the learning rate is $0.001$. It takes less than $30$ minutes to finish the a complete training and evaluation cycle for each vertical with one NVIDIA V100 GPU.

\begin{figure}[h]
\centering
  \includegraphics[width=.85\linewidth]{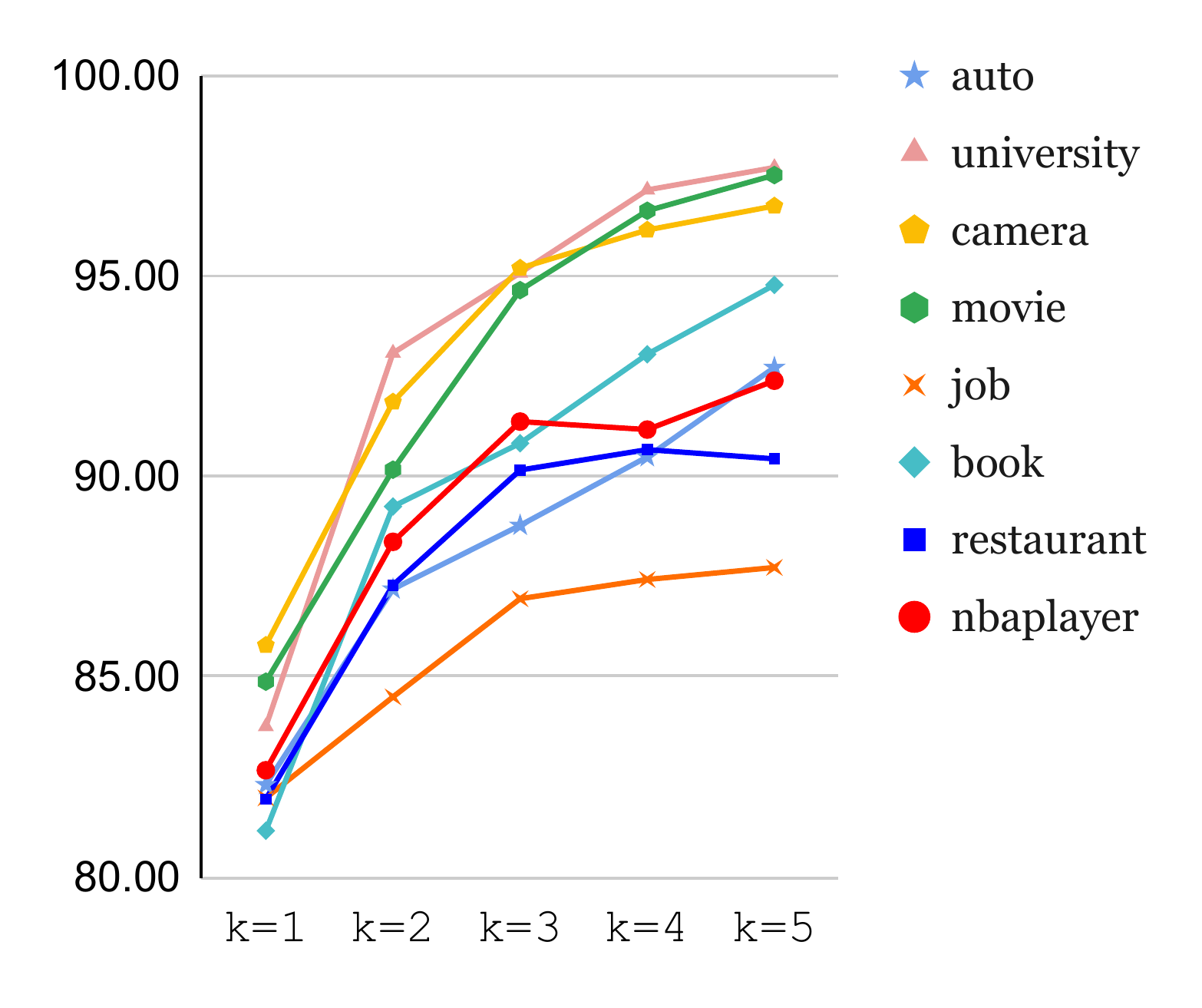}
  \caption{Comparing the extraction performance (F1 score) of different numbers of seed sites $k=\{1,2,3,4,5\}$ per vertical.}
  \label{fig:verticals}
    \vspace{-10pt}
\end{figure}

\subsection{Baseline Models}
We compare against several baselines:

\noindent\textbf{Stacked Skews Model (\texttt{SSM}).} \texttt{SSM}~\cite{carlson2008bootstrapping} utilizes expensive hand-crafted features and tree alignment algorithms to align the unseen web pages with seed web pages. This method does not require visual rendering features, which is the same as our model.

\noindent\textbf{Rendering-feature Model (\texttt{Render-full}). } 
\texttt{Render-full}~\cite{hao2011one} employs visual features to express the distances between node blocks rendered with the web browser. Visual distances are proven a good feature to encode the neighboring relationships among nodes~\cite{lockard2020zeroshotceres} but this method requires the time-consuming rendering process and needs extra memory space to save the images, CSS, and JavaScripts that can easily be out-of-date. 
In specific, \texttt{Render-full} employs a sophisticated heuristic algorithm to compute the visual distances, which gives the best performance~\cite{hao2011one}, compared to other variants \texttt{Render-PL} and \texttt{Render-IP}.

\noindent\textbf{Relational Neural Model (\texttt{FreeDOM-X}). }
\texttt{FreeDOM} leverages a relational neural network to encode features such as the relative distance and text semantics. This method is composed of two stages. The first stage model (\texttt{FreeDOM-NL}) learns a dense representation for each DOM tree node via node-level classification. The relational neural network in the second stage (\texttt{FreeDOM-Full}) claims to capture the distance and semantic relatedness 
between pairs of nodes in the DOM trees. This two-stage model does not rely on visual features but is hard to be deployed in practice. Besides, only modeling the relatedness between pairs of nodes neglects the rich structural information in the tree such as the friend circles. We compare with both \texttt{FreeDOM-NL} and \texttt{FreeDOM-Full} because the single-stage \texttt{FreeDOM-NL} is closer to our model and \texttt{FreeDOM-Full} achieves the state-of-the-art experimental results.

\begin{table}[h]
    \centering
    \resizebox{\linewidth}{!}{
    \begin{tabular}{|c|p{2.3em}|p{2.3em}|p{2.3em}|p{2.3em}|p{2.3em}|}
    \hline
        Model $\backslash$ \#Seed Sites & $k=1$ &  $k=2$ &  $k=3$ &  $k=4$ &  $k=5$\\
        \hline
        \hline
        \texttt{SSM} & 63.00 & 64.50 & 69.20 & 71.90 & 74.10\\
        \texttt{Render-Full} & \textbf{84.30} & 86.00 & 86.80 & 88.40 & 88.60 \\
        \texttt{FreeDOM-NL}  & 72.52 & 81.33 & 86.44 & 88.55 & 90.28 \\
        \texttt{FreeDOM-Full} & 82.32 & 86.36 & 90.49 & 91.29 & 92.56 \\
        \textbf{\modelname} & 83.06 & \textbf{88.96} & \textbf{91.63} & \textbf{92.84} & \textbf{93.75}\\
    \hline
    \end{tabular}
    }
    \vspace{10pt}
    \caption{Comparing the extraction performance (F1 score) of five baseline models to our method \modelname using different numbers of seed sites $k=\{1,2,3,4,5\}$. Each value in the table is computed from the average over 8 verticals and 10 permutations of seed websites per vertical (80 experiments in total).}
    \label{tab:main}
    \vspace{-15pt}
\end{table}

\subsection{Intra-vertical Few-shot Extraction Results}

Table~\ref{tab:main} shows the overall comparisons between our model \modelname and all four baselines using different numbers of seed websites. 
Our model achieves a slightly worse performance when $k=1$ while largely outperforms \texttt{Render-Full} when $k=\{2,3,4,5\}$. We can conclude that the delicately crafted visual features can capture more patterns in the scenario where extremely small training data exists. However, they are not as transferable as the rich semantic features extracted from our simplified DOM trees as $k$ increases. Our method also consistently outperforms the state-of-the-art method \texttt{FreeDOM-Full} (an average lift of 1.44\% over all the $k$'s) and achieves a 3.47\%-10.54\% improvement from the single-stage approach, \texttt{FreeDOM-NL}, per F1 score. 

We plot the detailed performance of \modelname on different verticals in figure~\ref{fig:verticals}. In general, the performance is improved as $k$ increases. This is not surprising because more training data obtain better coverage of all possible instances. we also observe that the rate of performance growth slows down and sometimes the F1 scores of some verticals (e.g. \textit{nbaplayer} and \textit{restaurant}) even fluctuates as more data join the training process (i.e. as $k$ increases). We think the reason is that the model becomes more robust and less new knowledge can be transferred from annotated websites to unseen websites in these verticals. 

\subsection{Ablation Study}
In Figure~\ref{fig:ablation}, we demonstrate an ablation study on different features of \modelname, including discrete features and friend circle features. We find that both sets of features improve the attribute extraction performance dramatically. For instance, the friend circle features lift up the F1 score of \textit{nbaplayer} vertical from $82.18\%$ to $91.37\%$ and the discrete features increase the performance on \textit{book} vertical by $8.51\%$. However, \textit{restaurant} is a special case where the result drops when we employ either of the two feature sets. We believe the node texts in some attribute values such as \textit{name} and \textit{address} are distinguishable enough and adding more features just brings more noise to the classification. 
This is also corroborated by Figure~\ref{fig:attributes}, which explains the detailed performance change when adding the friend circle features per attribute. We observe that the improvement on \textit{height} of \textit{nbaplayer} is significant. The nodes containing \textit{height} value always share a similar pattern xx-yy\footnote{For instance, NBA player Kobe Bryant's height (6-6) has the same value as his shooting record (6-6) in one game. It is impossible to distinguish two nodes by the text.} with some other nodes on the same page. 
With the friend circle features, we find that \textit{weight} is always a friend node of \textit{height}, which makes \textit{height} distinguishable from other nodes with similar text patterns.  

Another interesting ablation study is done with an additional sequence modeling layer\footnote{We utilize the Transformer~\cite{vaswani2017attention} as the sequence modeling layer. LSTM can be an alternative.} which is commonly applied to sequence labeling tasks such as named entity recognition on plain text~\cite{lample2016neural,yan2019tener}. 
We first obtain a sequence of node embeddings before the MLP classifier where all the nodes are from one web page. Then a new representation can be achieved from the sequence model for each node. The same classifier is used to predict the attribute type with the updated node representation.
As shown in Figure~\ref{fig:ablation} (marked as ``w/ Node Sequence Modeling''), the additional sequence modeling layer fails to optimize the node representations for all the verticals especially those with more variable nodes such as \textit{nbaplayer} and \textit{job}. We suppose that the information from all other DOM tree nodes can be selectively attended to the current node with such mechanism, which however introduces more noise than useful knowledge. This further proves the importance of utilizing the structures in the simplified DOM trees to eliminate the noise from distant and irrelevant nodes.


\begin{table}[h]
    \centering
    \resizebox{\linewidth}{!}{
    \begin{tabular}{|p{13em}|p{3em}|p{8.5em}|}
    \hline
        Embedding Approach & F1 & Performance Change\\
        \hline
        \hline
        GloVe Embedding Trainable & \textbf{91.63} & 0\\
        GloVe Embedding Fixed  &  91.25 & -0.38\\
        Randomized Word Embedding & 89.66 & -1.97\\
        Contextualied Embedding &  81.83 & -9.80\\
        
    \hline
    \end{tabular}
    }
    \vspace{10pt}
    \caption{Comparing different word embedding approaches when $k=3$.}
    \label{tab:embedding}
    \vspace{-10pt}
\end{table}

We also compare the different embedding approaches for encoding textual features. As shown in Table~\ref{tab:embedding}, we conduct experiments to test the randomized word embedding, fixed GloVe word embedding, and trainable GloVe word embedding. In the trainable setting, we can continue to optimize the parameters in the embedding layer which is initialized from GloVe and it gets the best performance. We think a specific ``web-language'' model can serve the web information extraction tasks better. As contextualized language models develop nowadays, we also try the BERT~\cite{devlin2018bert}\footnote{We choose BERT without loss of generality. It can be replaced by its alternatives like ELMo~\cite{peters2018deep} or XLNet~\cite{yang2019xlnet}.} to generate the contextualized embeddings but it decreases the performance by 9.8\%. It is not surprising because the context in each node is very limited\footnote{On average, each variable node contains only 2-5 words in different verticals.} and the huge size of parameters (110M in BERT-BASE) for fine-tuning can easily cause an over-fitting problem. 

\begin{figure}[h]
\centering
  \includegraphics[width=.9\linewidth]{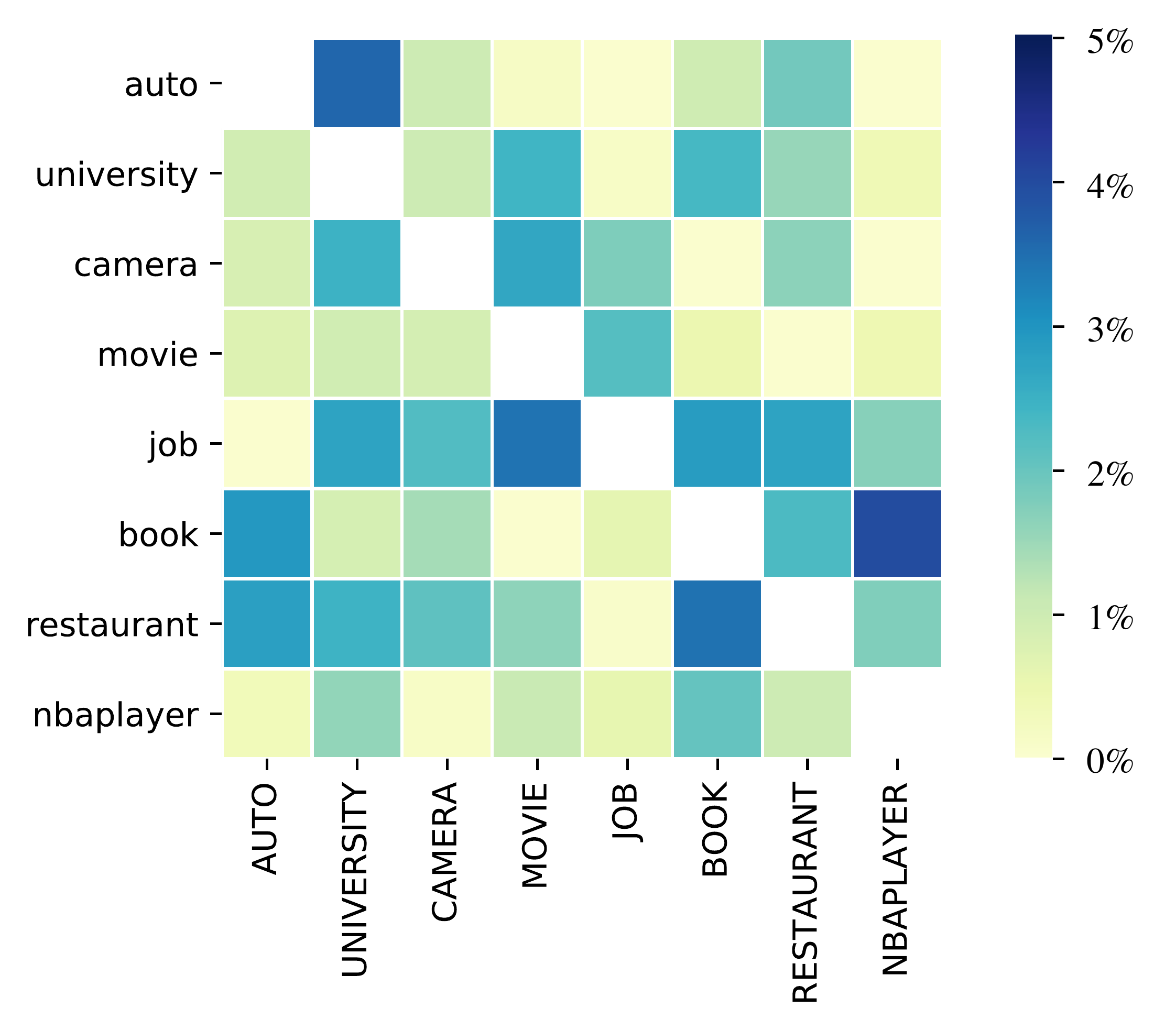}
  \vspace{-10pt}
  \caption{Heatmap denoting the performance lifts per F1 score from the out-of-domain knowledge. In specific, we learn a transferable model with verticals in upper case (columns). Then we finetune the model and predict on the verticals in lower case (rows). }
  \vspace{-10pt}
  \label{fig:cross}
\end{figure}

\subsection{Cross-vertical Few-shot Extraction Results}
We plot a heatmap in Figure~\ref{fig:cross} to denote the performance lifts from the out-of-domain knowledge. In specific, each entry in the heatmap relates to a pair of verticals, where the vertical in the upper case is used as the out-of-domain knowledge while the vertical in the lower case is used to train and test the model. 
We do not plot the scores in the diagonal because every vertical cannot serve as its out-of-domain resource.   
One interesting observation is that this heatmap is roughly symmetric with respect to the diagonal, which demonstrates a mutual relationship between pairs of verticals. For instance, \textit{job} and \textit{movie}, \textit{book} and \textit{nbaplayer}, \textit{restaurant} and \textit{book} can all significantly improve the extraction performance for each other, while \textit{auto} and \textit{job}, \textit{camera} and \textit{nbaplayer} seem to be irrelevant to each other. We show the performance of each vertical achieved by using the most helpful vertical's  out-of-domain knowledge in Figure~\ref{fig:ablation}. We achieve the highest average F1 score 93\% over all the verticals ($k=3$).

\section{Related Work} \label{sec:related}
\subsection{Web Information Extraction}
Web information extraction processes vast amount of unstructured or semi-structured contents from the web and has drawn a lot of attention from the data mining research community~\cite{chang2006survey,liu2018encyclopedia,baumgartner2001visual,popov2003towards,etzioni2008open}. 
Four broad categories of web information extraction tasks can be summarized. They are attribute (entity) extraction, relation extraction, composite extraction, and application-driven extraction.

\noindent \textbf{Attribute extraction} targets to identify named entity mentions such as book price, phone number, movie title from web documents. Though this task is intuitive to describe, the high-quality corpus annotation requires time-consuming human-crafted rules and dictionaries~\cite{lin2020freedom,hao2011one,carlson2008bootstrapping,pasupat2014zero}.  

\noindent \textbf{Relation extraction} associates pairs of named entities and identifies a pre-defined relationship between them. Closed relation extraction defines a closed set of relation types including a special type indicating "no relation" while open relation extraction conducts a binary classification of whether there exists a relationship between the two entities~\cite{augenstein2016distantly,zouaq2017assessment,quirk2016distant,lockard2018ceres,lockard2020zeroshotceres}.

\noindent \textbf{Composite extraction} aims to extract more complex concepts such as reviews, opinions, and sentiment mentions. Attribute and relation extractions can be integrated into the high-level workflow of composite extraction with other sub-modules like sentiment classification or aspect detection~\cite{das2007yahoo,chen2011empirical,song2010automatic,shandilya2009automatic,dave2003mining}.

\noindent \textbf{Application-driven extraction} includes a broad spectrum of application scenarios such as web representation learning, PDF information extraction using OCR techniques, anomaly detection of web-based attacks and so on~\cite{ramakrishnan2012layout,kruegel2005multi,vartouni2018anomaly,majumder2020representation,zhou2019learning,kocayusufoglu2019riser}.

\subsection{Attribute Extraction from Web Documents}

Attribute extraction serves as the fundamental task in the web information extraction pipelines and enables a wide range of downstream applications~\cite{bing2016unsupervised,dong2014knowledge,wu2018fonduer,wang2019multi}. However, there still exists a huge room to develop attribute extraction methods of high accuracy and strong transferability.
Traditional approaches~\cite{azir2017wrapper,kushmerick1997wrapper,muslea1999hierarchical,zheng2007joint,soderland1999learning,chang2001iepad,zhai2005web} either reply on analyzing the templates that are used to build the web pages or leverage unsupervised models to tackle the problem. However, they neglect the rich semantics of the attribute values and require considerable human efforts for annotations, thus failing to be generalizable to unseen websites.
Some recent methods~\cite{hao2011one,carlson2008bootstrapping} believe utilizing visual features generated from the web page rendering process can enable the model to extract attributes from new websites. Nevertheless, it is time-consuming to build visual features and space-unfriendly to store the necessary images, CSS, JavaScript files that are prone to be out-of-date. 
In this paper, we aim to construct a transferable model to extract attributes from unseen websites without using any visual features.

\section{Conclusion} \label{sec:conclusion}
In this paper, we propose a simple but effective method, \modelname, that simplifies the DOM trees to extract informative and transferable knowledge for the attribute extraction task. 
We build a rich representation for each DOM tree node without using any visual features. 
Extensive experiments show that \modelname significantly outperforms the SOTA method by 1.44\% on the F1 score and utilizing out-of-domain knowledge further improves the performance by 1.37\%.
We will open-source the implementations to facilitate further researches in the web data mining community.

\bibliographystyle{abbrv}
\bibliography{www21}

\end{document}